\documentclass[sigconf]{acmart}
\AtBeginDocument{%
  \providecommand\BibTeX{{%
    \normalfont B\kern-0.5em{\scshape i\kern-0.25em b}\kern-0.8em\TeX}}}

\copyrightyear{2024}
\acmYear{2024}
\setcopyright{rightsretained}
\acmConference[HRI '24 Companion]{Companion of the 2024 ACM/IEEE International Conference on Human-Robot Interaction}{March 11--14, 2024}{Boulder, CO, USA}
\acmBooktitle{Companion of the 2024 ACM/IEEE International Conference on Human-Robot Interaction (HRI '24 Companion), March 11--14, 2024, Boulder, CO, USA}
\acmDOI{10.1145/3610978.3641085}
\acmISBN{979-8-4007-0323-2/24/03}

\usepackage{setspace}
\usepackage[uppercase]{titlesec}
\titlespacing\section{0pt}{2pt plus 2pt minus 2pt}{0pt plus 2pt minus 2pt}
\titlespacing\subsection{0pt}{2pt plus 2pt minus 2pt}{0pt plus 2pt minus 2pt}
\titlespacing\subsubsection{0pt}{2pt plus 3pt minus 2pt}{0pt plus 2pt minus 2pt}

\begin{document}

\title{An Adaptable, Safe, and Portable Robot-Assisted Feeding System}





\author{Ethan K. Gordon}
\authornote{Equal contribution, alphabetical order}
\email{ekgordon@cs.washington.edu}
\affiliation{%
  \institution{University of Washington}
  \city{Seattle}
  \state{Washington}
  \country{USA}
}

\author{Rajat Kumar Jenamani}
\authornotemark[1]
\email{rj277@cornell.edu}
\affiliation{%
  \institution{Cornell University}
  \city{Ithaca}
  \state{New York}
  \country{USA}
}

\author{Amal Nanavati}
\authornotemark[1]
\email{amaln@cs.washington.edu}
\affiliation{%
  \institution{University of Washington}
  \city{Seattle}
  \state{Washington}
  \country{USA}
}

\author{Ziang Liu}
\author{Daniel Stabile}
\author{Xilai Dai}
\author{Tapomayukh Bhattacharjee}
\authornotemark[2]
\affiliation{%
  \institution{Cornell University}
  \city{Ithaca}
  \state{New York}
  \country{USA}
}

\author{Tyler Schrenk}
\author{Jonathan Ko}
\affiliation{%
    \institution{Researcher}
    \city{Seattle}
  \state{Washington}
  \country{USA}
}

\author{Haya Bolotski}
\author {Raida Karim}
\author{Atharva Kashyap}
\author{Bernie Hao Zhu}
\author{Taylor Kessler Faulkner}
\author{Siddhartha S. Srinivasa}
\authornote{Equal advising}
\affiliation{%
  \institution{University of Washington}
  \city{Seattle}
  \state{Washington}
  \country{USA}
}

\renewcommand{\shortauthors}{Ethan K. Gordon et al.}

\begin{abstract}
We demonstrate a robot-assisted feeding system that enables people with mobility impairments to feed themselves. Our system design embodies Safety, Portability, and User Control, with comprehensive full-stack safety checks, the ability to be mounted on and powered by any powered wheelchair, and a custom web-app allowing care-recipients to leverage their own assistive devices for robot control. For bite acquisition, we leverage multi-modal online learning to tractably adapt to unseen food types. For bite transfer, we leverage real-time mouth perception and interaction-aware control. Co-designed with community researchers, our system has been validated through multiple end-user studies.

\end{abstract}





\settopmatter{printacmref=true}
\maketitle
\section{Motivation and System Overview}

In the United States alone, at least 1.8 million people require assistance to eat \cite{theis2019one}. This requirement can result in feelings of shame and dependence among care recipients and places a significant burden on caregivers \cite{nanavati2023design}, especially because feeding is among the most time-consuming activity of daily living \cite{chio2006caregiver}. Robot-assisted feeding systems offer a promising solution for improving the quality of life for care recipients and alleviating caregiver burden. 

In this demonstration, we introduce a robot-assisted feeding system, collaboratively designed with end-users for real-world deployment. The system exemplifies the design principles of \emph{Safety}, \emph{Portability}, and \emph{User Control} in its hardware and software. Through a web app interface, the user controls the execution of bite acquisition (Section \ref{sec:acquisition}), which involves picking up food from a plate, and bite transfer (Section \ref{sec:transfer}), which involves delivering the food to the care-recipient's mouth. 
Our design has been informed by studies with end-users~\cite{bhattacharjee2019community, nanavati2023design} and ongoing work with community researchers with mobility impairments~\cite{nanavati2023design}. 
This system has been rigorously tested through end-user studies and deployments over several years \cite{gallenberger2019transfer, bhattacharjee2020moreautonomy, jenamani2024bitetransfer}, successfully feeding care-recipients with severe medical conditions such as Multiple Sclerosis, Spinal Cord Injury, Spinal Muscular Atrophy, and Arthrogryposis.
\begin{figure*}
    \centering
    \includegraphics[width=0.88\linewidth]{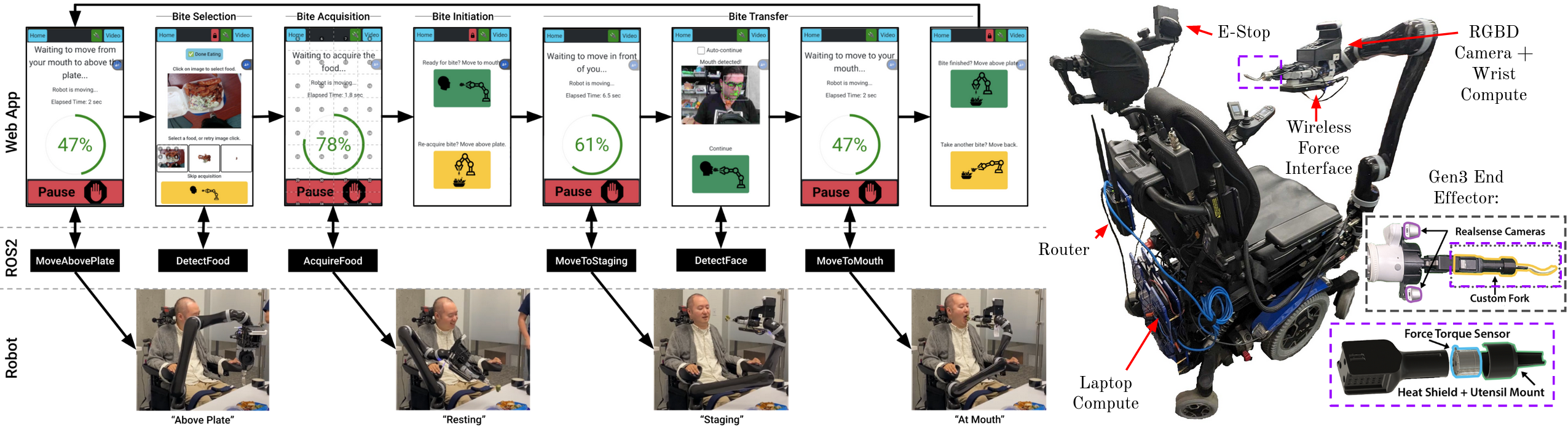}
    \caption{\small \emph{(Left)} Diagram of the system logic. The user drives the app, which calls an API on the robot computer. \emph{(Right)} Hardware system for both the Gen2 and Gen3 base. No external wires are needed.}
    \label{fig:system}
    \vspace{-0.5cm}
\end{figure*}

\subsection{\NoCaseChange{Hardware}}
\label{sec:hardware}
The system has been designed for 2 robot arms, the Kinova Gen2 (JACO) and the Kinova Gen3. Both have 6 degrees of freedom and can be powered directly from the user's wheelchair's battery. The JACO has been modified with an eye-in-hand system that still allows for continuous rotation of the wrist joint. The Gen3 has has been modified with a second camera. An Nvidia Jetson Nano developer kit mounted on the wrist manages an Intel Realsense D415 RGBD camera. When the feeding mode is active, the robot gripper holds onto a custom, 3d-printed fork assembly. The tines of the fork are attached to a 6-DOF ATI Nano25 force-torque transducer. The sensor is powered and read by ATI's battery-powered Wireless F/T. Finally, the primary compute is a Lenovo Legion 5 laptop with an Nvidia RTX 3060 6GB GPU, and the primary networking component is a Cradlepoint IBR900 router, both of which are mounted on the back of the user's wheelchair. Both systems can be connected directly to the 24V DC power provided on most wheelchairs, with the laptop specifically drawing 65W over USB-C.
The laptop in turn connects to the robot over USB and to a standard accessibility button over a 3.5mm aux cable for emergency stop. Links to exact hardware specifications can be found on our websites\footnote{\url{https://robotfeeding.io} and \url{https://emprise.cs.cornell.edu/bitetransfer}}.

\subsection{\NoCaseChange{Software}}
\label{sec:software}

Our software stack is built on ROS2 and the ros2-control framework\footnote{\url{https://control.ros.org}}. We have configured hardware driver plugins that can swap between the real robot, an IsaacSim, and a ``mock'' kinematics simulation. These plugins allow the entire system to operate both in simulation and in the real world without any other modifications. Above the hardware drivers is a collection of ROS2 controllers that handle trajectory execution, PID, and compliant control. Non-compliant velocity controllers implement ``force gating'', where execution will be aborted if the measured force or torque exceeds a configurable threshold. Higher-level functions use MoveIt2.

Individual robot commands are logically organized into behavior trees using \texttt{py\_trees}. Some trees handle multi-part complex actions: e.g. AcquireFood, which computes the food and approach reference frames, executes an element of our action schema (Sec.~\ref{sec:acquisition}), and returns the robot to the resting position. Other trees handle simpler actions: e.g. MoveAbovePlate, which executes a single MoveIt2 planning and execution call.

Our primary user interface is a React app that can be accessed via the browser of any phone, tablet, or computer. Many people with mobility impairments already utilize a variety of assistive technologies to interact with computers and mobile devices, such as a sip-n-puff, chin joystick, voice interface, or head-activated buttons. By operating as a web app on these devices, the feeding app can integrate with all of these systems without requiring additional configuration. Further, the app is designed to provide users high levels of control over and transparency into the system, to align with user preferences~\cite{nanavati2023design} and empower them to use the system without researcher intervention. Unlike previous iterations of the feeding system, the app represents the ``seat'' of the logic for our system. The finite state machine driving the app utilizes ROS2 action servers, services, and topics to initiate, monitor, and preempt robot actions, read from the sensors, write system parameters, and generally dictate the logical flow of the system (Fig.~\ref{fig:system}). The only elements that bypass the app control are the emergency stop and other safety components. This enables the user to have full control to start, stop, and modify system execution at any time.

\section{Design and Technical Details}

\subsection{\NoCaseChange{Design Principle: Safety}}
\label{sec:safety}

The robot-assisted feeding system should never cause harm to its user. This is particularly salient for feeding given the intimacy of the task and the potential danger of the tools involved. To achieve this, we've developed our system with four distinct layers of safety, each designed to protect end-users during the feeding process: 

\textbf{Compliant Hardware.} For in-mouth transfer, we use silicone utensils attached to a custom-designed utensil holder with a deliberately engineered mechanical weak point. Should forces exceeding a certain threshold be applied to the utensil's tip, the weak point will break. This is crucial for user safety; if too much force is applied while the utensil is in the user's mouth, it will break and only the silicone tip will remain in the mouth, stopping all physical interaction with the robot. In practice, our system has never required the utensil to break during inside-mouth bite transfers.

\textbf{Compliant Control.} We utilize a hard force threshold during non-transfer, velocity-controlled robot trajectories to stop all motion when an unexpected contact occurs. During the delicate inside-mouth transfer, we switch to torque control (with safe torque limits) and utilize a self-developed compliant control module, ensuring safety during physical interactions. Even incidental contact between the fork and 
mouth is only mildly uncomfortable.


\textbf{Software Anomaly Detection.} We verify hard safety constraints with a ``watchdog'' system: invariants are checked in a single node and an ``all-clear'' message is published frequently to the rest of the system. This minimizes the code that needs to be manually verified and simplifies the logic of other components: if an all-clear message has not been received in a certain period, shut down. For bite transfer, we also make use of explicit anomaly detection, such as  filtering outliers in mouth perception by leveraging a "likely range'' of head movements captured \emph{a priori}, and detecting and adapting to involuntary movement (muscle spasms) by the care-recipient. Our methods prioritizes safety: we prefer false positives (mistakenly detecting a spasm) over false negatives (missing a real spasm).

\textbf{Emergency Intervention.} Two safety invariants checked by the ``watchdog'' include the connectivity of the force sensor and an emergency stop button placed near the care recipient's head that they can click. When either fails, the robot controllers are immediately killed. Additionally, for cases where the care recipient cannot move their head, an experimenter sits beside the robot with a hardware switch at hand that can immediately cut robot power.

\subsection{\NoCaseChange{Design Principle: Portability}}
\label{sec:portabilty}

The robot-assisted feeding system should seamlessly fit into users' eating routines and environments. No mandatory wires connect the utensil to the rest of the robot or the wheelchair. Therefore, the robot has full range of motion with no danger of any wire snagging and causing damage, and a care recipient can have it put down the utensil after eating so they can use their robot arm for other tasks. Onboard compute---a laptop and router mounted to the back of the wheelchair---can be powered by any standard wheelchair battery. Our software functions without the need for reliable internet connectivity or external computation. 

\subsection{\NoCaseChange{Design Principle: User Control}}
\label{sec:user_interface}

The system should provide users control over the feeding process~\cite{nanavati2023design}. This is partially achieved by making the user interface the seat of logic (Sec.~\ref{sec:software}), controlling their path through the state machine. This ensures the user can pause at any time and decide what happens next, not just for safety, but also for comfort and convenience. We additionally emphasize customization: allowing the user to set the robot speed, bite transfer motion, interface mode, etc. This is in contrast to other systems like Obi \footnote{\url{https://meetobi.com/}}, which generally have a fixed configuration for a meal. Preferences can vary greatly between users and depend on the environment they are eating in, so users should be provided intuitive and expressive controls to customize the robot-assisted feeding system for their needs.

\begin{figure}[t]
    \centering
    \includegraphics[width=\linewidth]{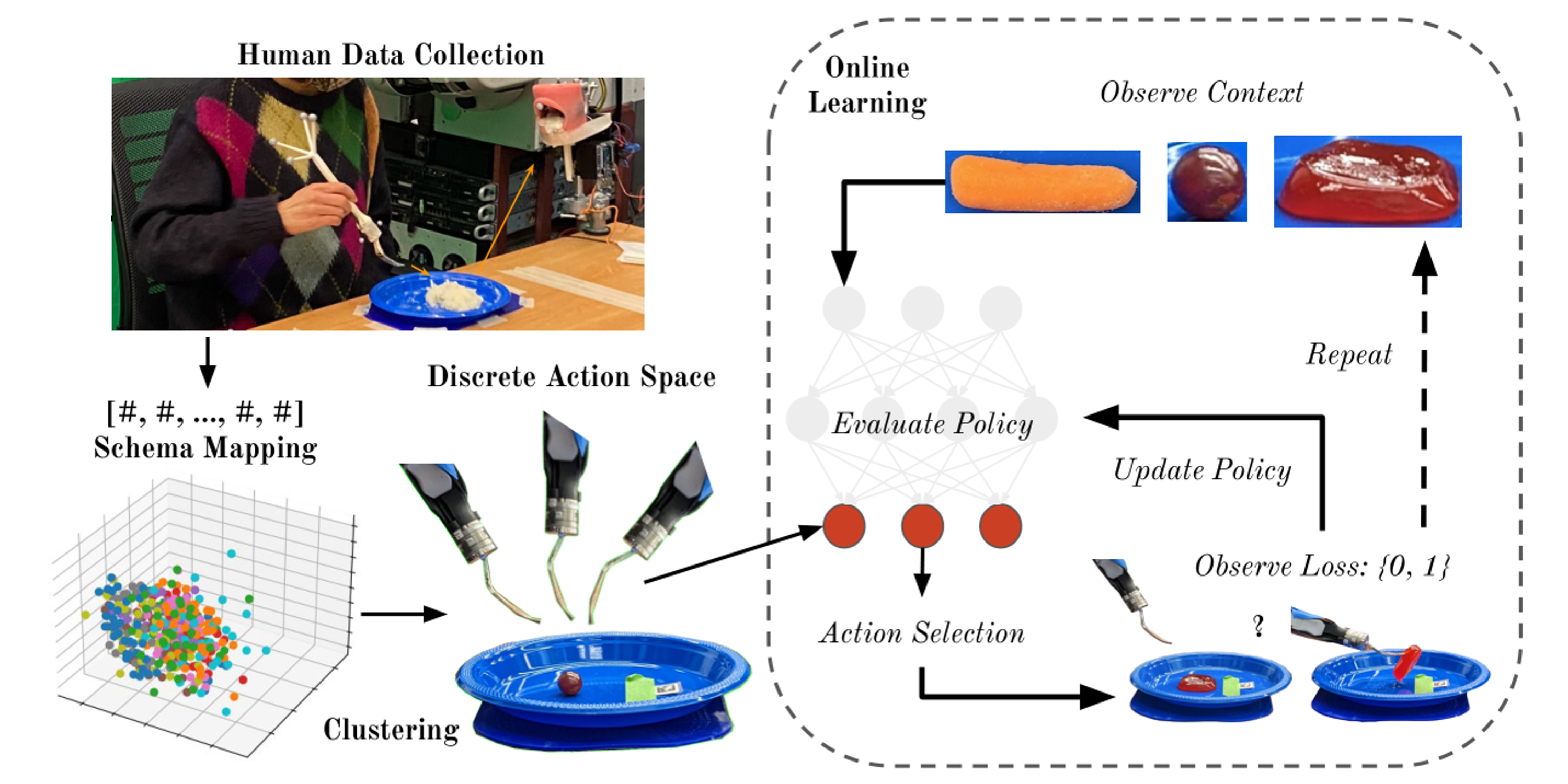}
    \vspace{-0.8cm}
    \caption{\small Bite acquisition: an action space derived from human data for online learning within a contextual bandit framework~\protect\cite{gordon2023schema}.}
    \label{fig:acquisition}
    \vspace{-0.5cm}
\end{figure}

\subsection{\NoCaseChange{Bite Acquisition}}
\label{sec:acquisition}

We demonstrate the tractably adaptable bite acquisition system published this year \cite{gordon2023schema}. The primary contribution is a 26-dimensional interpretable acquisition schema that encompasses a comprehensive taxonomy of human acquisition techniques \cite{bhattacharjee2019towards}, including skewering, scooping, and twirling. We collect $\sim500$ human acquisition trajectories to define an expert distribution in this space. Then, we use \emph{k-medioids} clustering as a form of spatially diverse sampling of that distribution. The result is a set of 11 discrete acquisition actions that features emergent behavior described qualitatively in previous work, such as in-food wiggling, as well as new effective behavior such as an extraction motion that tilts back so food falls onto the utensil. We utilize a contextual bandit framework augmented with haptic post hoc context\cite{gordon2021posthoc}, to achieve online adaptability. The robot will try different discrete actions with new food items and will figure out the best one over the course of 8-13 attempts.

\subsection{\NoCaseChange{Bite Transfer}}
\label{sec:transfer}

After picking up a food item from the plate, the robot transfers it to the care-recipient's mouth. For care-recipients able to lean forward, the robot positions the food at a user-customizable distance from their mouth, enabling them to bite off a stationary fork. For care-recipients who cannot lean forward, the robot places food directly in their mouths using a method \cite{jenamani2024bitetransfer} with two key components:

1. A real-time mouth perception method that combines inputs from multiple in-hand cameras to be robust to utensil occlusion. This method enables the robot to detect, adapt, and respond to both voluntary and involuntary head movements, and pause mid-approach if the user is not ready (talking / mouth closed).

2. A physical interaction-aware control method that uses multimodal sensing (visual + haptic) to discern the nature of the interaction and react accordingly. This enables the robot to navigate inside the user's mouth while avoiding incidental physical interactions, accommodate involuntary physical interactions, respond to in-mouth manipulation, and retract after an intentional bite.

\begin{figure}[t]
    \centering
    \includegraphics[width=\linewidth]{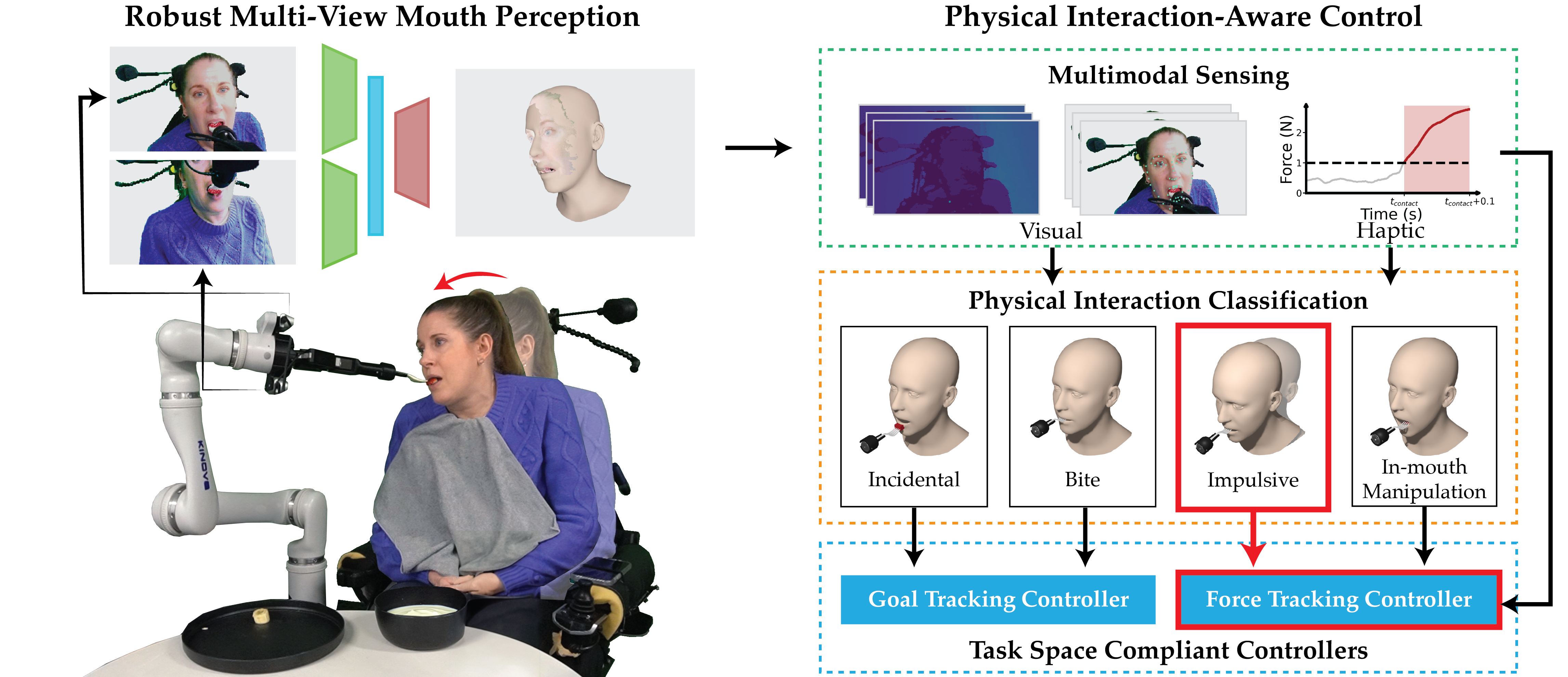}
    \vspace{-0.6cm}
    \caption{\small Bite transfer: multi-view mouth perception and physical interaction-aware control for an in-mouth hand-off~\protect\cite{jenamani2024bitetransfer}. }
    \label{fig:transfer}
    \vspace{-0.6cm}
\end{figure}


\vspace{-0.2cm}
{\small \paragraph{\bf \NoCaseChange{Funding Sources}} National Science Foundation NRI (\#2132848), CHS (\#2007011), IIS (\#2132846), CAREER (\#2238792), and GRFP (DGE-1762114), DARPA Under Contract (\#HR001120C0107), the Office of Naval Research (\#N00014-17-1-2617-P00004 and \#2022-016-01 UW), and Amazon}

\bibliographystyle{ACM-Reference-Format}
\bibliography{references}


\begin{thebibliography}{10}


\ifx \showCODEN    \undefined \def \showCODEN     #1{\unskip}     \fi
\ifx \showDOI      \undefined \def \showDOI       #1{#1}\fi
\ifx \showISBNx    \undefined \def \showISBNx     #1{\unskip}     \fi
\ifx \showISBNxiii \undefined \def \showISBNxiii  #1{\unskip}     \fi
\ifx \showISSN     \undefined \def \showISSN      #1{\unskip}     \fi
\ifx \showLCCN     \undefined \def \showLCCN      #1{\unskip}     \fi
\ifx \shownote     \undefined \def \shownote      #1{#1}          \fi
\ifx \showarticletitle \undefined \def \showarticletitle #1{#1}   \fi
\ifx \showURL      \undefined \def \showURL       {\relax}        \fi
\providecommand\bibfield[2]{#2}
\providecommand\bibinfo[2]{#2}
\providecommand\natexlab[1]{#1}
\providecommand\showeprint[2][]{arXiv:#2}

\bibitem[Bhattacharjee et~al\mbox{.}(2019a)]%
        {bhattacharjee2019community}
\bibfield{author}{\bibinfo{person}{Tapomayukh Bhattacharjee}, \bibinfo{person}{Maria~E. Cabrera}, \bibinfo{person}{Anat Caspi}, \bibinfo{person}{Maya Cakmak}, {and} \bibinfo{person}{Siddhartha~S. Srinivasa}.} \bibinfo{year}{2019}\natexlab{a}.
\newblock \showarticletitle{A Community-Centered Design Framework for Robot-Assisted Feeding Systems}. In \bibinfo{booktitle}{\emph{Proceedings of the 21st International ACM SIGACCESS Conference on Computers and Accessibility}} (Pittsburgh, PA, USA) \emph{(\bibinfo{series}{ASSETS '19})}. \bibinfo{publisher}{Association for Computing Machinery}, \bibinfo{address}{New York, NY, USA}, \bibinfo{pages}{482–494}.
\newblock
\showISBNx{9781450366762}
\urldef\tempurl%
\url{https://doi.org/10.1145/3308561.3353803}
\showDOI{\tempurl}


\bibitem[Bhattacharjee et~al\mbox{.}(2020)]%
        {bhattacharjee2020moreautonomy}
\bibfield{author}{\bibinfo{person}{Tapomayukh Bhattacharjee}, \bibinfo{person}{Ethan~K Gordon}, \bibinfo{person}{Rosario Scalise}, \bibinfo{person}{Maria~E Cabrera}, \bibinfo{person}{Anat Caspi}, \bibinfo{person}{Maya Cakmak}, {and} \bibinfo{person}{Siddhartha~S Srinivasa}.} \bibinfo{year}{2020}\natexlab{}.
\newblock \showarticletitle{Is more autonomy always better? exploring preferences of users with mobility impairments in robot-assisted feeding}. In \bibinfo{booktitle}{\emph{2020 15th ACM/IEEE International Conference on Human-Robot Interaction (HRI)}}. IEEE, \bibinfo{pages}{181--190}.
\newblock


\bibitem[Bhattacharjee et~al\mbox{.}(2019b)]%
        {bhattacharjee2019towards}
\bibfield{author}{\bibinfo{person}{Tapomayukh Bhattacharjee}, \bibinfo{person}{Gilwoo Lee}, \bibinfo{person}{Hanjun Song}, {and} \bibinfo{person}{Siddhartha~S Srinivasa}.} \bibinfo{year}{2019}\natexlab{b}.
\newblock \showarticletitle{Towards robotic feeding: Role of haptics in fork-based food manipulation}.
\newblock \bibinfo{journal}{\emph{IEEE Robotics and Automation Letters}} \bibinfo{volume}{4}, \bibinfo{number}{2} (\bibinfo{year}{2019}), \bibinfo{pages}{1485--1492}.
\newblock


\bibitem[Chi{\`o} et~al\mbox{.}(2006)]%
        {chio2006caregiver}
\bibfield{author}{\bibinfo{person}{Adriano Chi{\`o}}, \bibinfo{person}{A Gauthier}, \bibinfo{person}{A Vignola}, \bibinfo{person}{Andrea Calvo}, \bibinfo{person}{Paolo Ghiglione}, \bibinfo{person}{Enrico Cavallo}, \bibinfo{person}{AA Terreni}, {and} \bibinfo{person}{Roberto Mutani}.} \bibinfo{year}{2006}\natexlab{}.
\newblock \showarticletitle{Caregiver time use in ALS}.
\newblock \bibinfo{journal}{\emph{Neurology}} \bibinfo{volume}{67}, \bibinfo{number}{5} (\bibinfo{year}{2006}), \bibinfo{pages}{902--904}.
\newblock


\bibitem[Gallenberger et~al\mbox{.}(2019)]%
        {gallenberger2019transfer}
\bibfield{author}{\bibinfo{person}{Daniel Gallenberger}, \bibinfo{person}{Tapomayukh Bhattacharjee}, \bibinfo{person}{Youngsun Kim}, {and} \bibinfo{person}{Siddhartha~S Srinivasa}.} \bibinfo{year}{2019}\natexlab{}.
\newblock \showarticletitle{Transfer depends on acquisition: Analyzing manipulation strategies for robotic feeding}. In \bibinfo{booktitle}{\emph{2019 14th ACM/IEEE International Conference on Human-Robot Interaction (HRI)}}. IEEE, \bibinfo{pages}{267--276}.
\newblock


\bibitem[Gordon et~al\mbox{.}(2023)]%
        {gordon2023schema}
\bibfield{author}{\bibinfo{person}{Ethan~Kroll Gordon}, \bibinfo{person}{Amal Nanavati}, \bibinfo{person}{Ramya Challa}, \bibinfo{person}{Bernie~Hao Zhu}, \bibinfo{person}{Taylor Annette~Kessler Faulkner}, {and} \bibinfo{person}{Siddhartha Srinivasa}.} \bibinfo{year}{2023}\natexlab{}.
\newblock \showarticletitle{Towards General Single-Utensil Food Acquisition with Human-Informed Actions}. In \bibinfo{booktitle}{\emph{7th Annual Conference on Robot Learning (CoRL)}}. \bibinfo{address}{Atlanta, GA, USA}.
\newblock
\urldef\tempurl%
\url{https://openreview.net/forum?id=UZpWSDA3tZJ}
\showURL{%
\tempurl}


\bibitem[Gordon et~al\mbox{.}(2021)]%
        {gordon2021posthoc}
\bibfield{author}{\bibinfo{person}{Ethan~K. Gordon}, \bibinfo{person}{Sumegh Roychowdhury}, \bibinfo{person}{Tapomayukh Bhattacharjee}, \bibinfo{person}{Kevin Jamieson}, {and} \bibinfo{person}{Siddhartha~S. Srinivasa}.} \bibinfo{year}{2021}\natexlab{}.
\newblock \showarticletitle{Leveraging Post Hoc Context for Faster Learning in Bandit Settings with Applications in Robot-Assisted Feeding}. In \bibinfo{booktitle}{\emph{2021 IEEE International Conference on Robotics and Automation (ICRA)}} (Xi'an, China). \bibinfo{publisher}{IEEE Press}, \bibinfo{pages}{10528–10535}.
\newblock
\urldef\tempurl%
\url{https://doi.org/10.1109/ICRA48506.2021.9561520}
\showDOI{\tempurl}


\bibitem[Jenamani et~al\mbox{.}(2024)]%
        {jenamani2024bitetransfer}
\bibfield{author}{\bibinfo{person}{R.~K. Jenamani}, \bibinfo{person}{D. Stabile}, \bibinfo{person}{Z. Liu}, \bibinfo{person}{A. Anwar}, \bibinfo{person}{K. Dimitropoulou}, {and} \bibinfo{person}{T. Bhattacharjee}.} \bibinfo{year}{2024}\natexlab{}.
\newblock \showarticletitle{Feel the Bite: Robot-Assisted Inside-Mouth Bite Transfer using Robust Mouth Perception and Physical Interaction-Aware Control}. In \bibinfo{booktitle}{\emph{2024 19th ACM/IEEE International Conference on Human-Robot Interaction (HRI)}}.
\newblock


\bibitem[Nanavati et~al\mbox{.}(2023)]%
        {nanavati2023design}
\bibfield{author}{\bibinfo{person}{Amal Nanavati}, \bibinfo{person}{Patricia Alves-Oliveira}, \bibinfo{person}{Tyler Schrenk}, \bibinfo{person}{Ethan~K. Gordon}, \bibinfo{person}{Maya Cakmak}, {and} \bibinfo{person}{Siddhartha~S. Srinivasa}.} \bibinfo{year}{2023}\natexlab{}.
\newblock \showarticletitle{Design Principles for Robot-Assisted Feeding in Social Contexts}. In \bibinfo{booktitle}{\emph{Proceedings of the 2023 ACM/IEEE International Conference on Human-Robot Interaction}} (Stockholm, Sweden) \emph{(\bibinfo{series}{HRI '23})}. \bibinfo{publisher}{Association for Computing Machinery}, \bibinfo{address}{New York, NY, USA}, \bibinfo{pages}{24–33}.
\newblock
\showISBNx{9781450399647}
\urldef\tempurl%
\url{https://doi.org/10.1145/3568162.3576988}
\showDOI{\tempurl}


\bibitem[Theis et~al\mbox{.}(2019)]%
        {theis2019one}
\bibfield{author}{\bibinfo{person}{Kristina~A Theis}, \bibinfo{person}{Amy Steinweg}, \bibinfo{person}{Charles~G Helmick}, \bibinfo{person}{Elizabeth Courtney-Long}, \bibinfo{person}{Julie~A Bolen}, {and} \bibinfo{person}{Robin Lee}.} \bibinfo{year}{2019}\natexlab{}.
\newblock \showarticletitle{Which one? What kind? How many? Types, causes, and prevalence of disability among US adults}.
\newblock \bibinfo{journal}{\emph{Disability and health journal}} \bibinfo{volume}{12}, \bibinfo{number}{3} (\bibinfo{year}{2019}), \bibinfo{pages}{411--421}.
\newblock


\end{thebibliography}

\end{document}